\documentclass{article}

\usepackage{arxiv}
\usepackage{amsmath}
\usepackage{amssymb}
\usepackage[utf8]{inputenc} % allow utf-8 input
\usepackage[T1]{fontenc}    % use 8-bit T1 fonts
\usepackage{hyperref}       % hyperlinks
\usepackage{url}            % simple URL typesetting
\usepackage{float}
\usepackage{booktabs}       % professional-quality tables
\usepackage{amsfonts}       % blackboard math symbols
\usepackage{nicefrac}       % compact symbols for 1/2, etc.
\usepackage{microtype}      % microtypography
\usepackage{lipsum}
\usepackage{graphicx}
\graphicspath{ {./images/} }

\title{CAK: Emergent Audio Effects from Minimal Deep Learning}

\author{
 Austin Rockman\\
  \texttt{austin@gloame.ai} \\
}

\begin{document}
\maketitle
\begin{abstract}
We demonstrate that a single 3×3 convolutional kernel can produce emergent audio effects when trained on 200 samples from a personalized corpus. We achieve this through two key techniques: (1) Conditioning Aware Kernels (CAK), where output = input + (learned\_pattern × control), with a soft-gate mechanism supporting identity preservation at zero control; and (2) AuGAN (Audit GAN), which reframes adversarial training from "is this real?" to "did you apply the requested value?" Rather than learning to generate or detect forgeries, our networks cooperate to verify control application, discovering unique transformations. The learned kernel exhibits a diagonal structure creating frequency-dependent temporal shifts that are capable of producing musical effects based on input characteristics. Our results show the potential of adversarial training to discover audio transformations from minimal data, enabling new approaches to effect design.
\end{abstract}

% keywords can be removed

\section{Introduction}
Generative AI has captured the world’s imagination by transforming deep learning from an analytical tool into a creative medium. While GANs and diffusion models enable visual art and music generation, the manipulation of existing audio remains primarily rooted in traditional signal processing. Mathematical DSP has given us powerful audio effects, from convolution reverbs to analog circuit models, but these rely on human insight to translate acoustic phenomena into equations. What if we could learn audio effects directly from sound itself? We bridge this gap with Conditioning Aware Kernels (CAK), a modulation technique that discovers transformations directly from data through adversarial training.

Our approach explores an alternate perspective: neural networks as simplification approximators. By constraining our model to find minimal viable solutions, we investigate whether sophisticated audio transformations can arise from 200 training samples and 11 learnable parameters (a single 3×3 kernel with bias and scale). This framework allows us to study what emerges when model capacity is deliberately limited to match realistic data constraints.

Human perception mirrors this efficiency. We learn to recognize complex patterns from a handful of experiences. Similarly, experienced audio engineers develop intuition for effects through limited but focused interaction. CAK captures this principle computationally: given a small corpus of audio with varying features, our system learns not just to reproduce but to discover and apply learned qualities across unseen inputs.

Traditional GANs (Goodfellow et al., 2014) pit generator against discriminator in a forgery detection contest. We propose an 'audit game' (AuGAN) where the discriminator must verify that the generator applied the user's control value to its learned features, whatever those features turn out to be. This structural shift, from deception to verification, enables the discriminator to guide the discovery of audio transformations. 

Through CAK, we investigate whether neural transformations require architectural complexity, or whether they can emerge from the interaction between minimal learned operations and the inherent richness of audio signals.

\section{Related Work}
\label{sec:headings}
\textbf{Neural Audio Synthesis and Effects}: WaveGAN (Donahue et al., 2018) and GANSynth (Engel et al., 2019) first showed that adversarial training can generate raw waveform audio using large datasets and models. DDSP (Engel et al., 2020) and RAVE (Caillon \& Esling, 2021) achieve high quality synthesis through compact architectures and strong inductive biases, enabling efficient training even with limited data. Unlike those works, CAK is not a standard generator; it learns effects from a small, personalized corpus.

\textbf{Conditioning Mechanisms}: Feature‑wise Linear Modulation (FiLM; Perez et al., 2018) conditions deep networks via channel wise affine transforms. In a 2019 retrospective, the authors note that FiLM often needs additional task‑specific inductive biases to remain data efficient (Perez et al., 2019 retrospective), a limitation we also observed when applying FiLM to complex conditioning vectors. Dynamic kernel methods such as CondConv (Yang et al., 2019) synthesize weights by mixing basis filters. CondConv effectively asks which kernel a layer should use; CAK instead detects a salient pattern and modulates its residual contribution by the user’s control value.

\textbf{Few-Shot Learning}: Audio few-shot work typically relies on meta‑learning (MAML, Finn et al., 2017; Prototypical Networks, Snell et al., 2017), which require access to large and diverse meta-training sets composed of many tasks. CAK operates in a more extreme regime: it learns directly from a single, 50‑minute corpus without episodic sampling or meta‑tasks.

\textbf{Emergent Complexity from Simple Rules}: Growing Isotropic Neural Cellular Automata (Mordvintsev, Randazzo, and Fouts, 2022) demonstrates similar principles in the visual domain, where simple local update rules produce complex emergent patterns. Like CAK, this work shows that behavioral diversity can arise from the interaction between minimal fixed rules and varying initial conditions, rather than from architectural complexity.

\textbf{Biological Inspiration}: Our emphasis on minimal, data‑efficient representations echoes the efficient coding hypothesis (Barlow, 1961) and sparse coding results in V1 (Olshausen \& Field, 1996) which suggest biological systems seek minimal representations. Feature Integration Theory (Treisman \& Gelade, 1980) likewise suggests that selective modulation of simple detectors can explain complex percepts, paralleling CAK’s single‑kernel modulation experiment.

\section{Method}

\subsection{Empirical Motivation}
Our initial approach followed established conditioning methods, using FiLM (Perez et al., 2018) with 24-dimensional control vectors encoding categorical tags, continuous DSP parameters, and perceptual attributes. This exhibited training instability. FiLM's affine transformations rely on deep networks to achieve complex modulation, making them potentially unsuitable for rich musical descriptors and small datasets. We do not claim that FiLM is fundamentally limited in this regard, only that we could not find the optimal solution for this problem using conventional modulation methods. 

This failure revealed an insight: complex control may not always require complex modulation. Through ablation, we discovered that our initially complex network consistently relied on a small subset of detected patterns, regardless of the control vector's dimensionality. This suggested a different approach: instead of learning how to modulate based on complex inputs, learn what patterns to detect, then simply scale them.

The audit game naturally enforces this sparsity. The discriminator must verify control values, incentivizing the generator to find distinct patterns. This led to CAK: a single learned detector whose output is scaled by a scalar control value. The shift from 24-dimensional modulation to 1-dimensional scaling did not reduce expressiveness; it revealed that one well-learned pattern could create effects through context-aware application.
\begin{figure}[H] % picture
    \centering
    \includegraphics[width=1\linewidth]{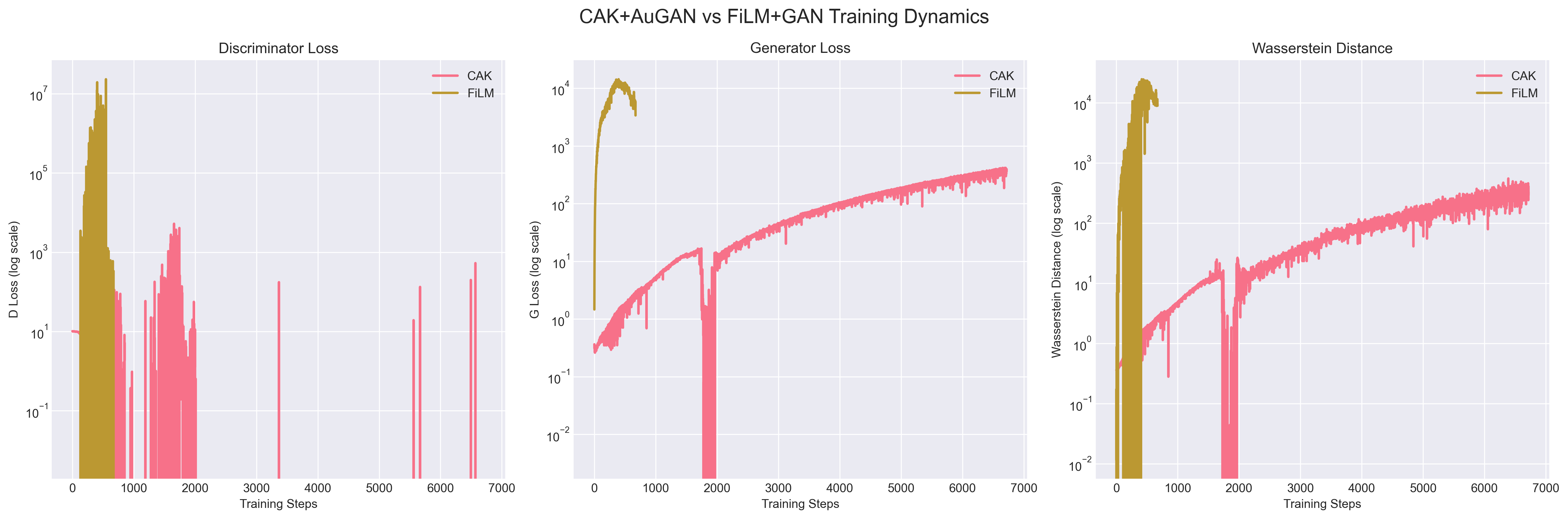}
\caption{\textit{Training dynamics comparison between FiLM and CAK architectures. Both networks were trained under identical conditions with no divergence mitigation strategies. FiLM-based conditioning with 24-dimensional control vectors exhibits instability, with discriminator and generator loss exceeding $10^4$ and training failure. In contrast, CAK achieves stable training on the same dataset, enabling the discovery of patterns despite its ultimate convergence to a single dominant feature.}}
\end{figure}

Instead of fighting against the results of the network, we theorized that dimensional collapse might contain a fascinating study of how neural networks process complex domains. We consider music to be infinitely analyzable. A single sample contains many possible interpretations. Collapse is usually treated as a failure to be mitigated through architectural tricks or regularization, and in some cases may be the correct motive, but perhaps we could design architectures that embrace, rather than resist, this tendency toward simplification.         

\subsection{CAK Architecture}
The core CAK operation implements a simple principle:
\[
\text{output = input + (learned\_pattern} \times \text{control)}
\]
Formally, this becomes:
\begin{equation}
y = x + (D(x) \times c \times \sigma(c) \times s)
\end{equation}
where:
\begin{itemize}
    \item $x \in \mathbb{R}^{F \times T}$ is the input magnitude spectrogram
    \item $D: \mathbb{R}^{F \times T} \rightarrow \mathbb{R}^{F \times T}$ is a learned $3 \times 3$ convolutional detector (same padding, with bias)
    \item $c \in \mathbb{R}$ is a per-example control scalar that broadcasts over the $F \times T$ dimensions
    \item $\sigma(c) = \text{sigmoid}((c - \tau) \times \text{temp})$ is a soft-gate function with:
    \begin{itemize}
        \item $\tau \in \mathbb{R}$: threshold (0.3 in our experiments, ensuring control values below 0.3 produce minimal activation)
        \item $\text{temp} \in \mathbb{R}$: temperature parameter (2 $\rightarrow$ 20 linear ramp during training); lower values create gradual transitions, higher values approach a hard cutoff at $\tau$
    \end{itemize}
    \item $s \in \mathbb{R}$ is a learned scale parameter that correlates with effect intensity
\end{itemize}
During training, each spectrogram is paired with a randomly sampled control value c. This scalar multiplies the detected patterns D(x) element-wise across the entire spectrogram, creating a simple yet effective modulation mechanism. The audit game ensures the network learns to detect patterns whose intensity scales meaningfully with c. We think of c as a continuous modulation knob where higher values produce proportionally stronger effects. While we use random sampling in this work, the framework supports experiments using c to encode possible semantic attributes. 

The soft-gate $\sigma(c)$ provides smooth onset based on the control value, which also directly scales the effect through multiplication. We choose multiplication as it preserves sonic character while scaling intensity, aligning with human auditory perception, which responds to amplitude ratios rather than absolute differences. This is arguably the simplest scaling solution possible. The dual use of c ensures both proportional intensity and gated activation, while the residual path assists with transparency at zero control.

\textbf{Key Properties:}
\begin{enumerate}
    \item \textbf{Identity Preservation}: At $c = 0$, the residual term $c \cdot \sigma(c)$ is zero, so $y = x$ exactly; $\tau$ and temp shape only the onset for small $c$.
    \item \textbf{Additive Modulation}: Unlike multiplicative attention, additive residual modulation preserves the original signal pathway, reducing risk of information loss compared to multiplicative gating.
    \item \textbf{Shared Detection}: The shared detector $D$ is updated jointly by generator and discriminator (critic) gradients; we did not freeze it in either branch.
\end{enumerate}

The soft-gate mechanism enables precise control. With threshold $\tau = 0.3$ and temperature annealing:
\begin{itemize}
    \item Control values below $\tau$ produce minimal effect
    \item Smooth transition around $\tau$ prevents discontinuities
    \item High final temperature creates sharp but differentiable gating
\end{itemize}

Unlike FiLM (per-channel affine $\gamma, \beta$ conditioned on the input) or CondConv/dynamic conv (per-input kernels via routing), CAK uses a single shared detector $D$ and a user-supplied scalar $c$ (with a soft gate) to scale a fixed residual.

\subsection{The AuGAN Framework}
Traditional GANs optimize $\min_G \max_D V(D,G)$ where $G$ tries to fool $D$. We reformulate this as AuGAN (Audit GAN), where both networks cooperate to verify control application:

\textbf{Generator Objective}: Apply transformations proportional to control value

\textbf{Discriminator Objective}: Verify if the correct control amount was applied

Crucially, both networks share the same detector $D$. This prevents the generator from learning arbitrary transformations, any pattern it uses must help the discriminator verify control values. AuGAN's cooperative dynamics promote:
\begin{enumerate}
    \item \textbf{Distinct Features}: Random patterns will not help verification
    \item \textbf{Proportional Application}: The transformation strength must scale consistently with the control value
    \item \textbf{Smooth Control}: The discriminator needs to distinguish nearby control values
\end{enumerate}

We implement AuGAN using WGAN-GP with additional terms to enforce control compliance. Following WGAN convention, we refer to the discriminator as the Critic ($C$):

\textbf{Discriminator Loss:}
\begin{equation}
L_C = -\mathbb{E}[C(x_{\text{real}}, c)] + \mathbb{E}[C(x_{\text{fake}}, c)] + \lambda_{\text{gp}} \cdot \text{GP} + \lambda_{\text{comp}} \cdot \mathbb{E}[V(x_{\text{fake}}, c)]
\end{equation}

\textbf{Generator Loss:}
\begin{equation}
L_G = -\mathbb{E}[C(x_{\text{fake}}, c)] + \lambda_{\text{comp}} \cdot \mathbb{E}[V(x_{\text{fake}}, c)] + \lambda_{\text{recon}} \cdot ||x_{\text{fake}} - x_{\text{real}}||_1 - \lambda_{\text{reg}} \cdot \mathbb{E}[\log(\epsilon + \text{mean}_{F,T}|D(x_{\text{in}})|)]
\end{equation}

\textbf{Pairing strategy:} We train on tuples $(x_{\text{in}}, x_{\text{real}}, c)$ of two types: (i) Identity pairs: $x_{\text{real}} = x_{\text{in}}$, $c = 0$ anchor the operator to the identity. (ii) Transformation pairs: $x_{\text{in}} = x_{\text{low}}$, $x_{\text{real}} = x_{\text{high}}$, $c = g(\text{high}) - g(\text{low})$ supply a target change and a control value. The L1 term pulls $G(x_{\text{in}}, c)$ toward $x_{\text{real}}$ while the compliance term enforces $m(G(x_{\text{in}}, c)) \approx c$; together with shared $D$, this rules out the trivial copy solution yet preserves identity at $c = 0$.

where:
\begin{itemize}
    \item $x_{\text{fake}} = G(x_{\text{in}}, c)$ is the generator output
    \item $x_{\text{real}}$ is the paired target from the dataset:
    \begin{itemize}
        \item For identity pairs: $x_{\text{real}} = x_{\text{in}}$ with $c = 0$
        \item For transformation pairs: $x_{\text{real}}$ is a different sample with $c$ encoding their relationship
    \end{itemize}
    \item $C(\cdot)$ outputs realness score and violation $V(\cdot,c)$ for control verification
    \item $V(x,c) = |\text{measured\_texture}(x) - c|$ where $\text{measured\_texture}(x) = \text{mean}(D(x))$ using the shared detector, we measure texture as a signed mean of $D(x)$; regularization uses $|D(x)|$ to avoid collapse
    \item GP is the gradient penalty for Lipschitz constraint computed on interpolations between $x_{\text{real}}$ and $x_{\text{fake}}$ with the same control value $c$
    \item $D(x_{\text{in}})$ in the regularization term refers to the shared detector patterns on input
    \item $\epsilon = 10^{-8}$ before the logarithm in the regularizer for numerical stability
    \item $\lambda_{\text{gp}} = 10.0$, $\lambda_{\text{comp}} = 2.0$, $\lambda_{\text{recon}} = 5.0$, $\lambda_{\text{reg}} = 0.01$
\end{itemize}

\section{Experiments}
\subsection{Experimental Setup}
We designed our experiments to reflect realistic artistic workflows. Musicians and sound designers typically work with curated personal collections rather than massive datasets. Our setup mirrors this reality:

\textbf{Dataset}: 200 fifteen-second audio segments derived from the author's musical corpus, representing the scale of material an artist might realistically collect and curate for a specific project. This includes varied timbral content mainly centered around electronic and electroacoustic music composition: synthesized textures, field recordings, and acoustic instrumentation.

\textbf{Preprocessing}: STFT with 2048-point FFT, 512 sample hop, 44.1 kHz sample rate, standard parameters for musical applications allowing the learned kernel to discover patterns directly from minimally processed spectrograms.

\textbf{Training}: 100 epochs on Apple M4 (48GB unified memory), completing in approximately 2 hours. Figure 2 shows stable training without divergence. 

Audio demonstrations, code, and our interactive GUI where users can test their own sounds are available at: \url{https://github.com/gloame-ai/cak-audio}.

\begin{figure}[H]
    \centering
    \includegraphics[width=1\linewidth]{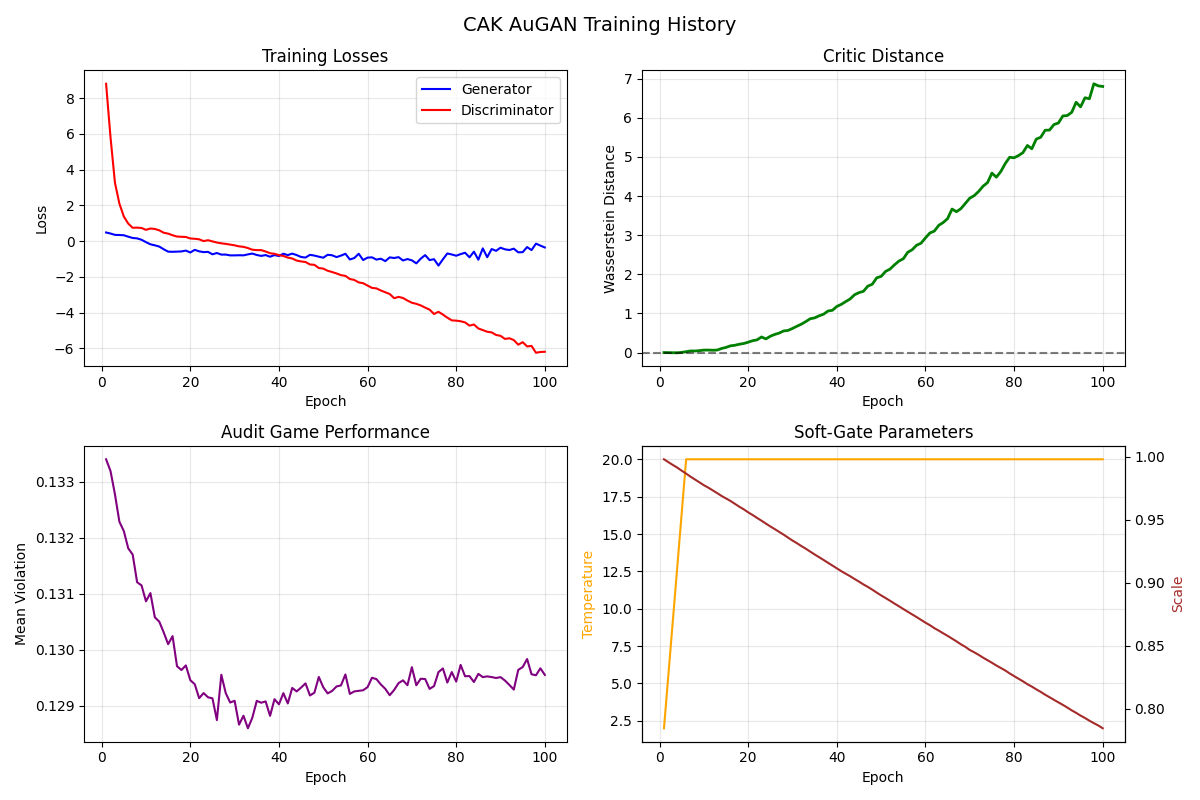}
    \caption{\textit{Training dynamics of CAK over 100 epochs using 200 15-second samples. Generator and discriminator losses show stable convergence. Increasing Wasserstein distance indicates healthy adversarial learning. Decreasing audit violations demonstrate successful effect control learning. Temperature annealing (orange) sharpens the soft-gate while the scale parameter (brown) adapts to optimal effect strength.} }
    \label{fig:enter-label}
\end{figure}

\subsection{Identity Preservation}
The identity constraint at control value zero is a calibration mechanism, ensuring that our learned transformation maintains ideal magnitude reconstruction when no modulation is desired. This helps prevent neural spectral coloration in bypass mode and forces the network to learn truly residual transformations. Identity preservation was tested on held out, diverse audio sources, with gate activation of 0.0025 on average at zero control and negligible magnitude difference (difference $< 10^{-9}$), confirming the soft-gate mechanism contributes to transparent pass-through. Naturally, various inputs will have different responses and we do not claim perfect unity gain at bypass. These results can be tested audibly in our GUI by simply processing a sample at a control value of 0.

\subsection{Emergent Behavior and Kernel Analysis}
The learned $3 \times 3$ detector kernel reveals how CAK achieves feature learning through a single pattern. Figure 3 shows the learned weights and their interpretable structure.
\begin{figure}[H]
    \centering
    \includegraphics[width=1\linewidth]{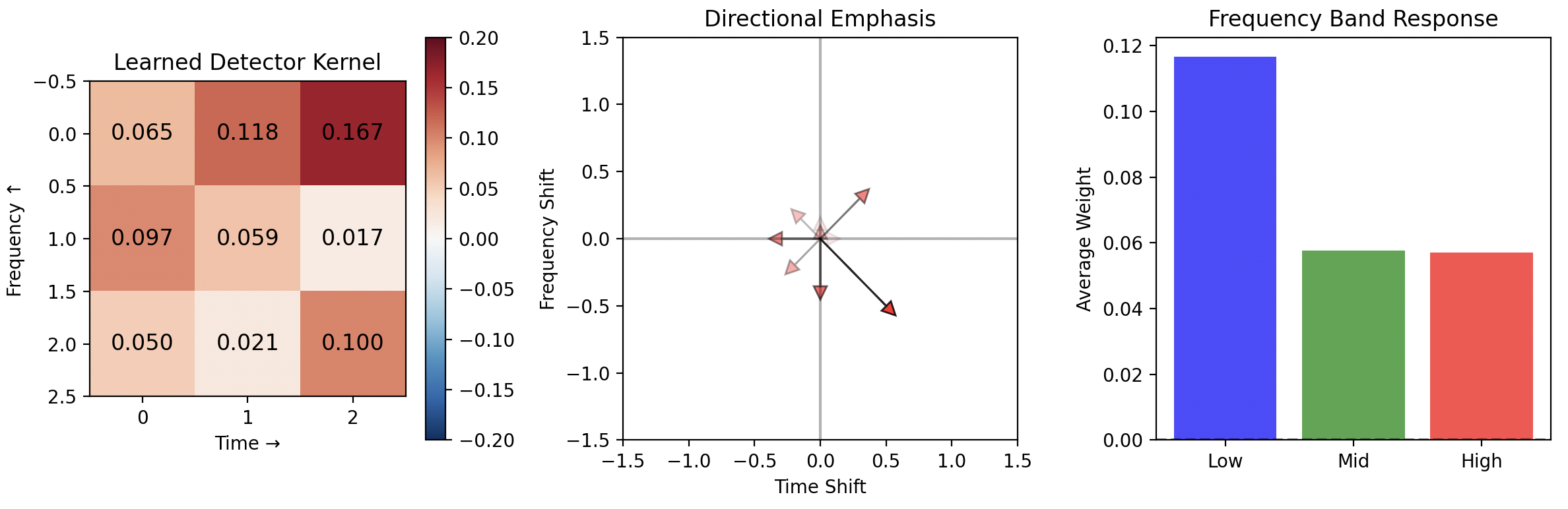}
    \caption{\textit{The learned detector kernel shows the asymmetric pattern that underlies CAK's behavior. During convolution with STFT magnitude, each weight indicates how strongly that time-frequency relationship contributes to the output. The rightward-biased pattern (high weights at positions [0,2], [0,1], and [2,2]) creates asymmetric time-frequency smoothing, emphasizing future time steps and producing spectral-temporal diffusion across the magnitude representation. The frequency band response (right) shows emergent selectivity, with stronger low-frequency weighting (0.115) despite no explicit frequency conditioning during training. This demonstrates how the CAK framework discovers both spectral and temporal patterns directly from data.}}
    \label{fig:enter-label}
\end{figure}

The learned transformation resists simple categorization, which we view as a fundamental characteristic of neural audio processing. Unlike traditional effects with clear design goals, CAK discovers patterns that produce varied perceptual results across different input types. This difficulty in defining the transformation using conventional audio terminology highlights both a limitation and a strength: while we cannot provide a traditional taxonomy, the effect represents a fascinating transformation learned from the data itself. This suggests future possibilities for audio processing that move beyond recreating known effects.

\section{Future Work}
Several directions merit exploration:

\textbf{Alternative Training Frameworks}: Although we trained CAK using adversarial dynamics, the architecture itself may be training agnostic. Investigating CAK within VAE frameworks or through direct supervised learning could reveal different emergent behaviors and potentially simpler training procedures.

\textbf{Semantic Control}: Our current approach learns effects from data without semantic labels. Incorporating dilated convolutions or attention mechanisms may enable targeting specific perceptual qualities (e.g., ``brightness,'' ``warmth'') while maintaining our minimal parameter philosophy.  

\textbf{Architectural Extensions}: Stacking multiple CAK layers with varying kernel sizes may capture multi-scale patterns. Additionally, frequency-band-specific CAK modules may enable surgical audio manipulation by applying different learned transformations to isolated spectral regions and recombining them for complex, structured effects. We are currently focusing further research efforts in this direction. 

\textbf{Cross-Domain Applications}: The principle of learning minimal patterns that interact with input characteristics may extend beyond audio. Investigating CAK on image or video data could validate whether this emergence phenomenon generalizes across modalities.

\section{Conclusion}

By constraining neural audio processing to a single 3×3 convolutional kernel, we have demonstrated that compelling audio effects can emerge from just 200 training samples and 11 learnable parameters. In this experiment, CAK acts as a learned texture neuron, a transformation that complements rather than replaces the tradition of hand-designed effects. Where traditional DSP encodes human understanding, CAK lets the data itself reveal what attributes of the spectra align with user control values.

The emergent behaviors observed in CAK, from frequency-dependent modulation to adaptive spectral enhancement, arise from the interaction between minimal structure and input diversity rather than model capacity. However, the learned transformation resists simple categorization using traditional audio terminology, and further work is needed to understand the relationship between training corpora characteristics and the resulting effects. Extending CAK to test multi-scale architectures and formulations or semantic control remains an open area for investigation.

We hypothesize that neural networks, when presented with minimally expressive structures and appropriate training dynamics, approximate complex behaviors by discovering simplicity rather than accumulating complexity. CAK validates this hypothesis, opening new directions for both audio processing and neural architecture design.

\section*{Acknowledgments}
We would like to acknowledge Roopam Garg of Gloame AI for implementing the demonstration GUI, contributions toward identity preservation logic, and providing iterative feedback.

\section*{References}
Barlow, H. (1961). Possible principles underlying the transformation of sensory messages. In \textit{Sensory communication} (pp. 217-234). MIT Press.

Caillon, A., \& Esling, P. (2021). RAVE: A variational autoencoder for fast and high-quality neural audio synthesis. \textit{arXiv preprint arXiv:2111.05011}.

Donahue, C., McAuley, J., \& Puckette, M. (2018). Adversarial audio synthesis. In \textit{International Conference on Learning Representations}.

Engel, J., Agrawal, K. K., Chen, S., Gulrajani, I., Donahue, C., \& Roberts, A. (2019). GANSynth: Adversarial neural audio synthesis. In \textit{International Conference on Learning Representations}.

Engel, J., Hantrakul, L., Gu, C., \& Roberts, A. (2020). DDSP: Differentiable digital signal processing. In \textit{International Conference on Learning Representations}.

Finn, C., Abbeel, P., \& Levine, S. (2017). Model-agnostic meta-learning for fast adaptation of deep networks. In \textit{International Conference on Machine Learning} (pp. 1126-1135).

Goodfellow, I., Pouget-Abadie, J., Mirza, M., Xu, B., Warde-Farley, D., Ozair, S., Courville, A., \& Bengio, Y. (2014). Generative adversarial nets. In \textit{Advances in neural information processing systems} (pp. 2672-2680).

Mordvintsev, A., Randazzo, E., \& Fouts, C. (2022). Growing isotropic neural cellular automata. In \textit{Artificial Life Conference Proceedings} (pp. 1-8).

Olshausen, B. A., \& Field, D. J. (1996). Emergence of simple-cell receptive field properties by learning a sparse code for natural images. \textit{Nature}, \textit{381}(6583), 607-609.

Perez, E., Strub, F., De Vries, H., Dumoulin, V., \& Courville, A. (2018). FiLM: Visual reasoning with a general conditioning layer. In \textit{Proceedings of the AAAI Conference on Artificial Intelligence}.

Perez, E., Strub, F., De Vries, H., Dumoulin, V., \& Courville, A. (2019). FiLM: Visual reasoning with a general conditioning layer - A retrospective. In \textit{ML Retrospectives Workshop at NeurIPS 2019}.

Snell, J., Swersky, K., \& Zemel, R. (2017). Prototypical networks for few-shot learning. In \textit{Advances in neural information processing systems} (pp. 4077-4087).

Treisman, A. M., \& Gelade, G. (1980). A feature-integration theory of attention. \textit{Cognitive psychology}, \textit{12}(1), 97-136.

Yang, B., Bender, G., Le, Q. V., \& Ngiam, J. (2019). CondConv: Conditionally parameterized convolutions for efficient inference. In \textit{Advances in Neural Information Processing Systems} (pp. 1307-1318).

\end{document}